\documentclass{svmult}
  
\usepackage{mathptmx,hyperref}   
\usepackage{helvet} 
\usepackage{courier}
\usepackage{amsmath,amsfonts}
\usepackage{graphicx}
\usepackage[bottom]{footmisc}          
\usepackage{sidecap}
    
\usepackage{mathptmx}         
\usepackage{helvet}            
\usepackage{courier}            
\usepackage{type1cm}             
\usepackage{makeidx}             
\usepackage{graphicx}          
\usepackage{multicol}          
\usepackage[bottom]{footmisc}  
     
\usepackage{epsfig}     
\usepackage{psfrag,rotating}     
\usepackage{amssymb,floatflt,enumerate} 
\usepackage{amsmath,amscd,psfrag,leqno} 
\usepackage[mathscr]{eucal}  
\usepackage{shadethm}           
\usepackage{graphicx,subfigure}   
\usepackage{overpic,contour}       
\contourlength{0.3mm}

\newcommand{\dzl}[1]{\underline{#1}}

\DeclareMathOperator{\sgn}{sgn}
\DeclareMathOperator{\diag}{diag}

\makeindex

\begin{document}

\title*{Construction and deformation of P-hedra \\ using control polylines}

\author{Georg Nawratil}
\authorrunning{G. Nawratil}
\institute{
  Institute of Discrete Mathematics and Geometry \&  
	Center for Geometry and Computational Design, TU Wien, 
  \email{nawratil@geometrie.tuwien.ac.at}}

%
%
\maketitle

\abstract{
In the 19th International Symposium on Advances in Robot Kinematics the author introduced  
a novel class of continuous flexible discrete surfaces and mentioned that these 
so-called P-hedra (or P-nets) allow direct access to their spatial shapes by three control polylines. 
In this follow-up paper we study this intuitive method, which makes these flexible planar quad surfaces suitable for transformable design tasks by means of interactive tools. 
The construction of P-hedra from the control polylines can also be used for an efficient algorithmic computation of their isometric deformations. 
In addition we discuss flexion limits, bifurcation configurations, developable/flat-foldable pattern and tubular P-hedra. 
}

\keywords{rigid-foldability, planar quad-surface, bifurcation, flexion limits, P-hedral tubes}

\section{Introduction}\label{sec:intro}

A planar quad-surface (PQ-surface) is a plate-and-hinge structure made of quadrilateral panels connected by rotational joints in the combinatorics of a square grid.  
Such a surface is called {\it continuous flexible} (or {\it rigid-foldable} or {\it isometric deformable}) if it can be continuously transformed by a change of the dihedral angles only. 
It is well known that the rigid-foldability of PQ-surfaces
is not a property of the extrinsic geometry
but of the intrinsic one \cite{Izm17}, which is determined by the
corner angles of the planar quads. 
Nonetheless, certain classes of rigid-foldable 
PQ-surfaces, namely V-hedra (and their related surfaces \cite{voss}) and T-hedra originally introduced by 
Sauer and Graf \cite{graf}, allow for direct access to their spatial shape by control polylines. 

In Section \ref{sec:construction} we show that this also holds for P-hedra which makes them suitable for transformable design tasks using interactive tools like V-hedra and T-hedra (see \cite{voss,SNRT2021,KMN1} and the references therein). In Section \ref{sec:defomration} we give an algorithm for the isometric deformations within the class of P-hedra and discuss possible bifurcation configurations and flexion limits, respectively (cf.\ Remarks \ref{rem:no_bifur} and \ref{rem:bifur}). 
Moreover, in Section \ref{sec:final} we comment on developable/flat-foldable pattern, the construction of P-hedral tubes and future research.

\begin{figure}[t]
\begin{overpic}
    [width=38mm]{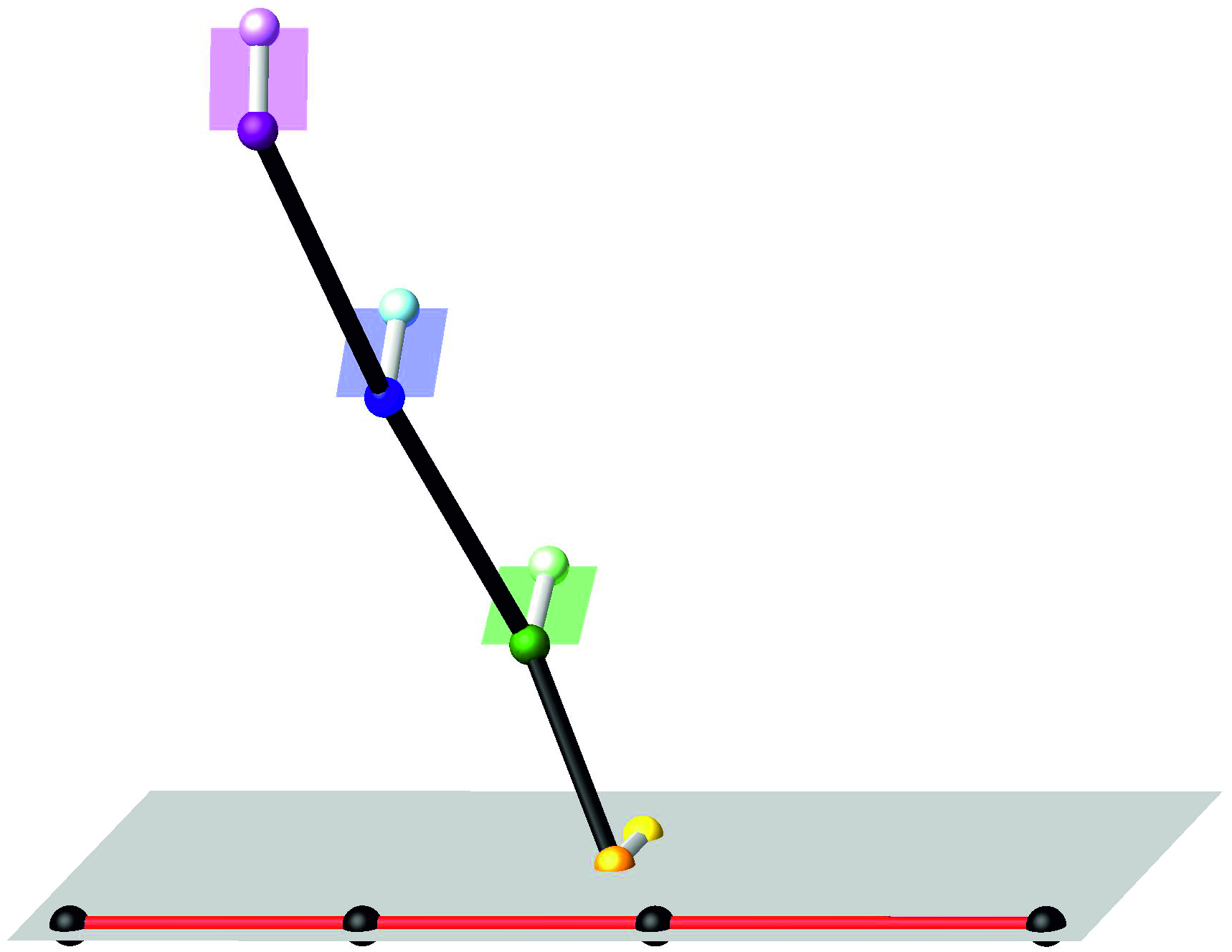}
    \put(37,7){\scriptsize\contour{white} {${V}_{0,0}$}}
    \put(31,22){\scriptsize\contour{white} {${V}_{1,0}$}}
    \put(8.5,63){\scriptsize\contour{white} {${V}_{m,0}$}}
    \put(50.5,12.5){\scriptsize\contour{white} {$D_0$}}
    \put(42.2,33.7){\scriptsize\contour{white} {$D_1$}}
    \put(18,77.5){\scriptsize\contour{white} {$D_m$}}
    \put(86,8){\scriptsize\contour{white} {$\pi_{0}$}}
    \put(48.5,26.5){\scriptsize\contour{white} {$\pi_{1}$}}
    \put(25.8,69.2){\scriptsize\contour{white} {$\pi_{m}$}}
    \put(26,-5.7){\scriptsize\contour{white} {$S_{0}$}}
    \put(2,-5.7){\scriptsize\contour{white} {$S^\pm_{1}$}}
    \put(82,-5.7){\scriptsize\contour{white} {$S_{n}$}}
  \end{overpic} 
\hfill
 \begin{overpic}
    [width=38mm]{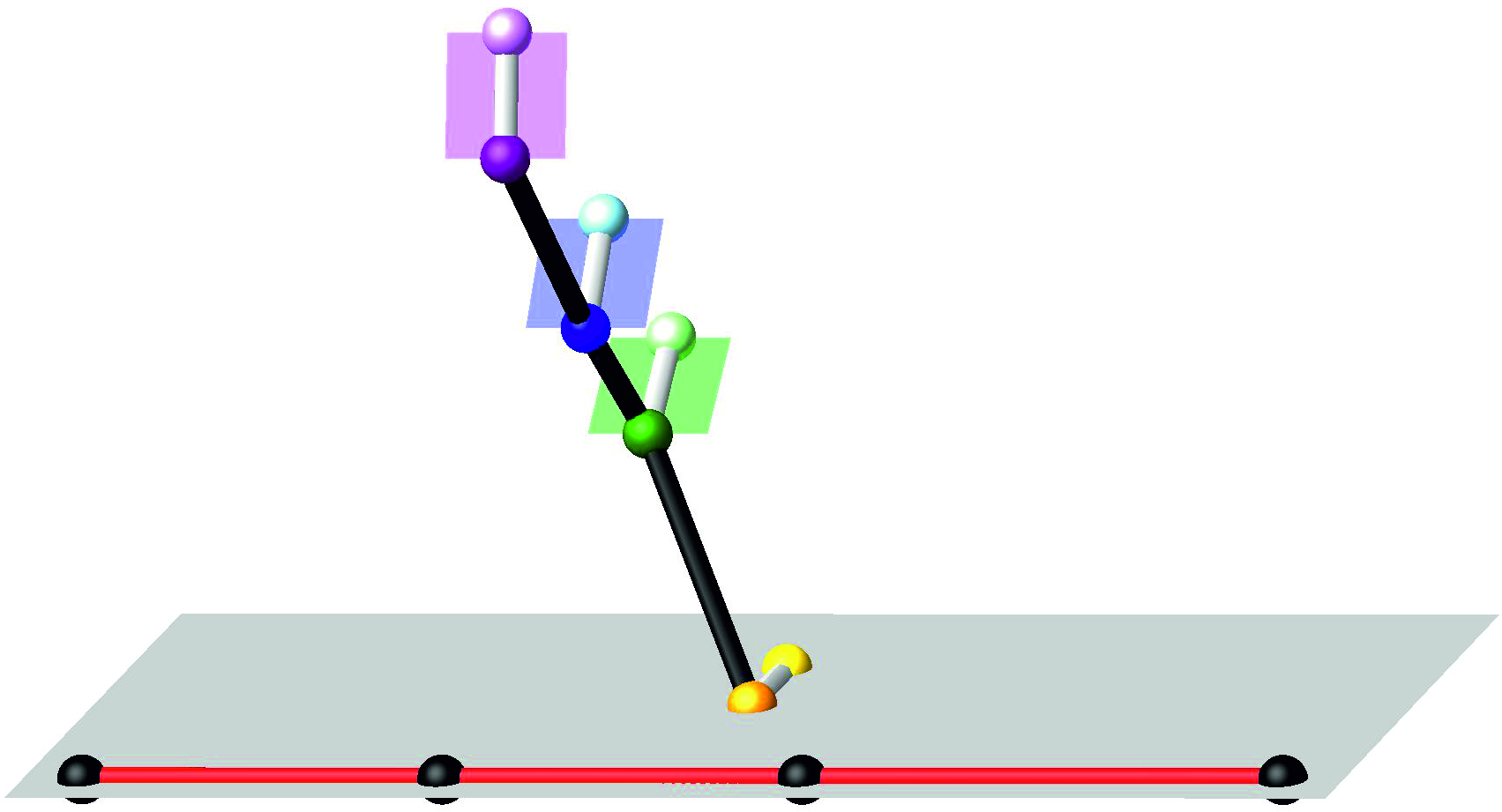}
     \put(37,7){\scriptsize\contour{white} {${V}^{\circ}_{0,0}$}}
    \put(31,22){\scriptsize\contour{white} {${V}^{\circ}_{1,0}$}}
    \put(21.5,41){\scriptsize\contour{white} {${V}^{\circ}_{m,0}$}}
    \put(50.5,12.5){\scriptsize\contour{white} {$D^{\circ}_0$}}
    \put(44,34){\scriptsize\contour{white} {$D^{\circ}_1$}}
    \put(31.5,54.5){\scriptsize\contour{white} {$D^{\circ}_m$}}
    \put(86,8){\scriptsize\contour{white} {$\pi^{\circ}_{0}$}}
    \put(48.5,26.5){\scriptsize\contour{white} {$\pi^{\circ}_{1}$}}
    \put(38,46){\scriptsize\contour{white} {$\pi^{\circ}_{m}$}}
    \put(26,-5.7){\scriptsize\contour{white} {$S_{0}$}}
    \put(2,-5.7){\scriptsize\contour{white} {$S^\pm_{1}$}}
    \put(82,-5.7){\scriptsize\contour{white} {$S_{n}$}}
  \end{overpic} 
\hfill	
\begin{overpic}
    [width=34mm]{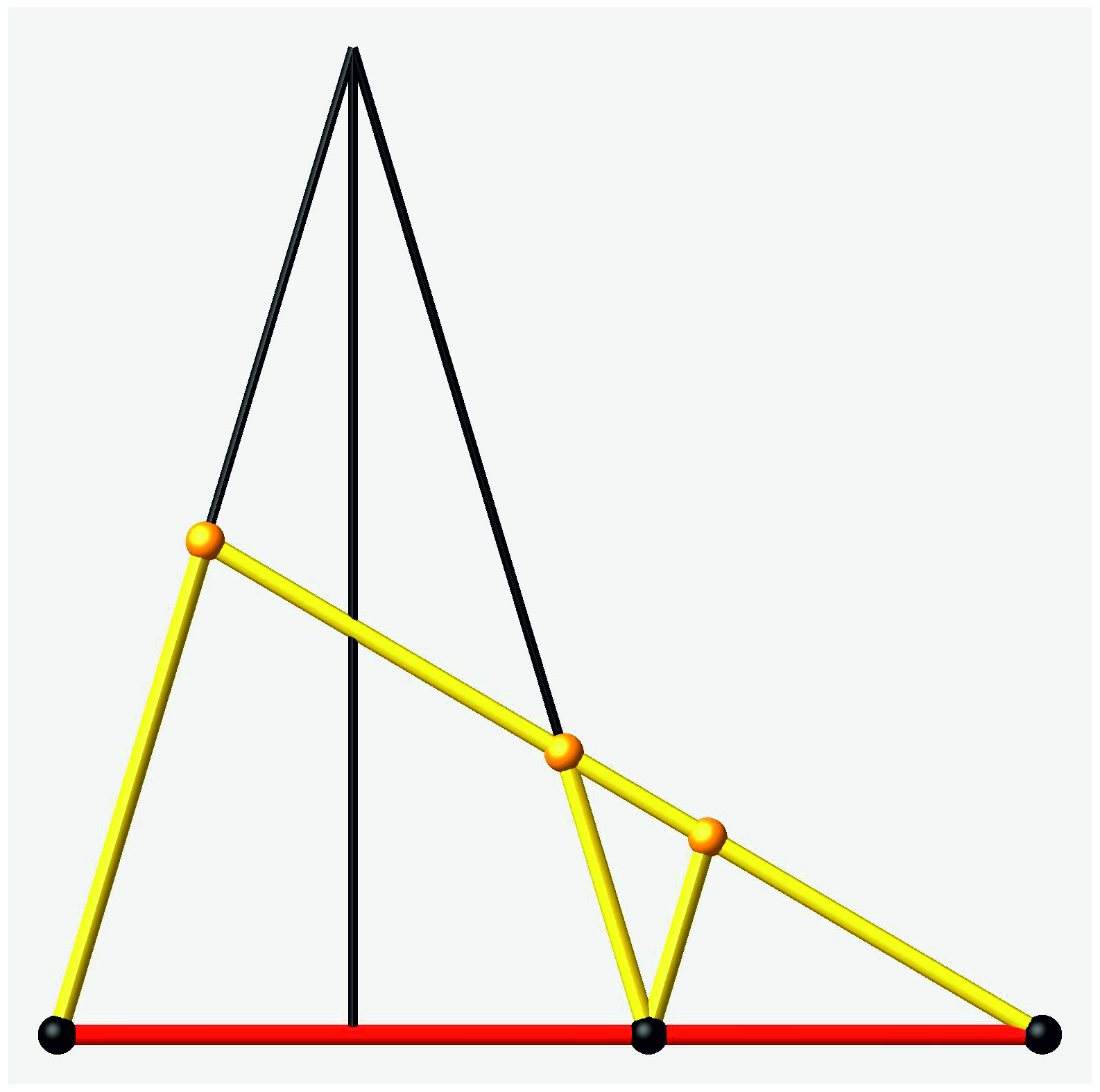}
    \put(3,-5.7){\scriptsize\contour{white} {$S_{j}$}}
    \put(88,-5.7){\scriptsize\contour{white} {$S^\pm_{j+1}$}}
    \put(53,-5.7){\scriptsize\contour{white} {$S_{j+2}$}}
    \put(7,50){\scriptsize\contour{white} {${V}^{\circ}_{i,j}$}}
    \put(52,34){\scriptsize\contour{white} {$\sigma^{-}({V}^{\circ}_{i,j})$}}
   \put(67,23){\scriptsize\contour{white} {$\sigma^{+}({V}^{\circ}_{i,j})$}}
   \put(27,22){\scriptsize\contour{white} {$q$}}
  \end{overpic} 
  \medskip
\hfill
\caption{(left) Input of a general P-hedron: trajectory polyline $V_{0,0}, V_{1,0}, \ldots ,V_{m,0}$, direction polyline $D_0, D_1, \ldots , D_m$ and apex polyline $S_0, S_1^\pm ,\ldots ,S_{n-1}^\pm, S_{n}$. 
(center) Input of the associated axial P-hedron: trajectory polyline $V^{\circ}_{0,0}, V^{\circ}_{1,0}, \ldots ,V^{\circ}_{m,0}$, direction polyline $D^{\circ}_0, D^{\circ}_1, \ldots , D^{\circ}_m$ and apex polyline $S_0, S_1^\pm ,\ldots ,S_{n-1}^\pm, S_{n}$. 
(right) Illustration of the two constructions related to the sign of $S_{j+1}^\pm$.}
  \label{fig1}
\end{figure}
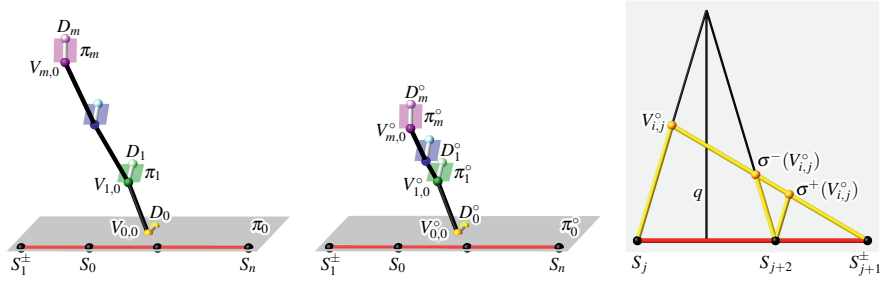  

\section{Reconstruction of general P-hedra from three control polylines}
\label{sec:construction}
In the following we describe how a general P-hedron can be reconstructed from three polylines illustrated in Fig.\ \ref{fig1}-left. 

Given is a so-called trajectory polyline $V_{0,0}, V_{1,0}, \ldots ,V_{m,0}$. 
Moreover, we have a polyline formed by direction points $D_0, D_1, \ldots , D_m$ with $V_{i,0}\neq D_i$. These points cannot be selected arbitrarily but the vector $\overrightarrow{V_{i,0}D_i}$ has to 
be orthogonal to a fixed direction, which can be assumed to be the $z$-direction of the fixed frame; i.e.\ $\langle D_i-V_{i,0}, z \rangle =0$ where $\langle .,.\rangle$ denotes the standard scalar product. 
As the lengths of the vectors $\overrightarrow{V_{i,0}D_i}$ are not of relevance, they can be assumed as unit-vectors. 

The plane through $V_{i,0}$ and $D_i$ which is parallel to the $z$-direction is called profile plane $\pi_i$. 
Moreover, we assume that no two consecutive profile planes are 
neither identical\footnote{\label{fn:bifur} $\pi_i = \pi_{i+1}$ and $V_{i,0}\neq V_{i+1,0}$ implies a bifurcation configuration (cf.\ Remark \ref{rem:bifur}).} 
nor parallel\footnote{\label{fn:trans}  $\pi_i \parallel \pi_{i+1}$ implies some special treatment as we get a translational surface-strip (cf.\ Section \ref{sec:final}).}. 
Without loss of generality (w.l.o.g.) we can assume that the $z$-axis equals the intersection line of $\pi_0$ and $\pi_1$. 
Moreover, we can assume w.l.o.g.\ that $\pi_0$ equals the $xz$-plane and that $V_{0,0}$ is located on the positive $x$-axis\footnote{We assume that $V_{0,0}$ differs from the origin, in order 
to avoid a degenerated case.}.
Beside the trajectory polyline and the direction polyline, we can select finite\footnote{\label{fn:special} As mentioned in \cite{nawratil}, $S_i$ can also be an ideal point but we do not consider this special case here.} points $S_0, S_1 ,\ldots , S_n$ on the $z$-axis; i.e.\ $S_i=(0,0,z_i)^T$.  
In order to avoid degenerated cases we assume that three consecutive vertices $S_j$, $S_{j+1}$ and $S_{j+2}$ are always pairwise distinct.  
In addition, we assign to each of the points $S_1,\ldots, S_{n-1}$ either a plus or a minus sign\footnote{The $\pm$ assignment can be done arbitrarily with exception of the case mentioned in Remark \ref{rem:sissor}.}, 
which results in the sequence  $S_0, S_1^\pm ,\ldots ,S_{n-1}^\pm, S_{n}$. We denote this polyline as apex polyline. 
This nomenclature becomes clear at the end of this section.

Finally, we assume that the trajectory polyline is not contained in the $xy$-plane because then the P-hedron belongs also to the class of T-hedra, which are already well studied and understood (cf.\ \cite{graf,SNRT2021,sauer,arvin}).

\begin{figure}[t]
\begin{overpic}
    [width=38mm]{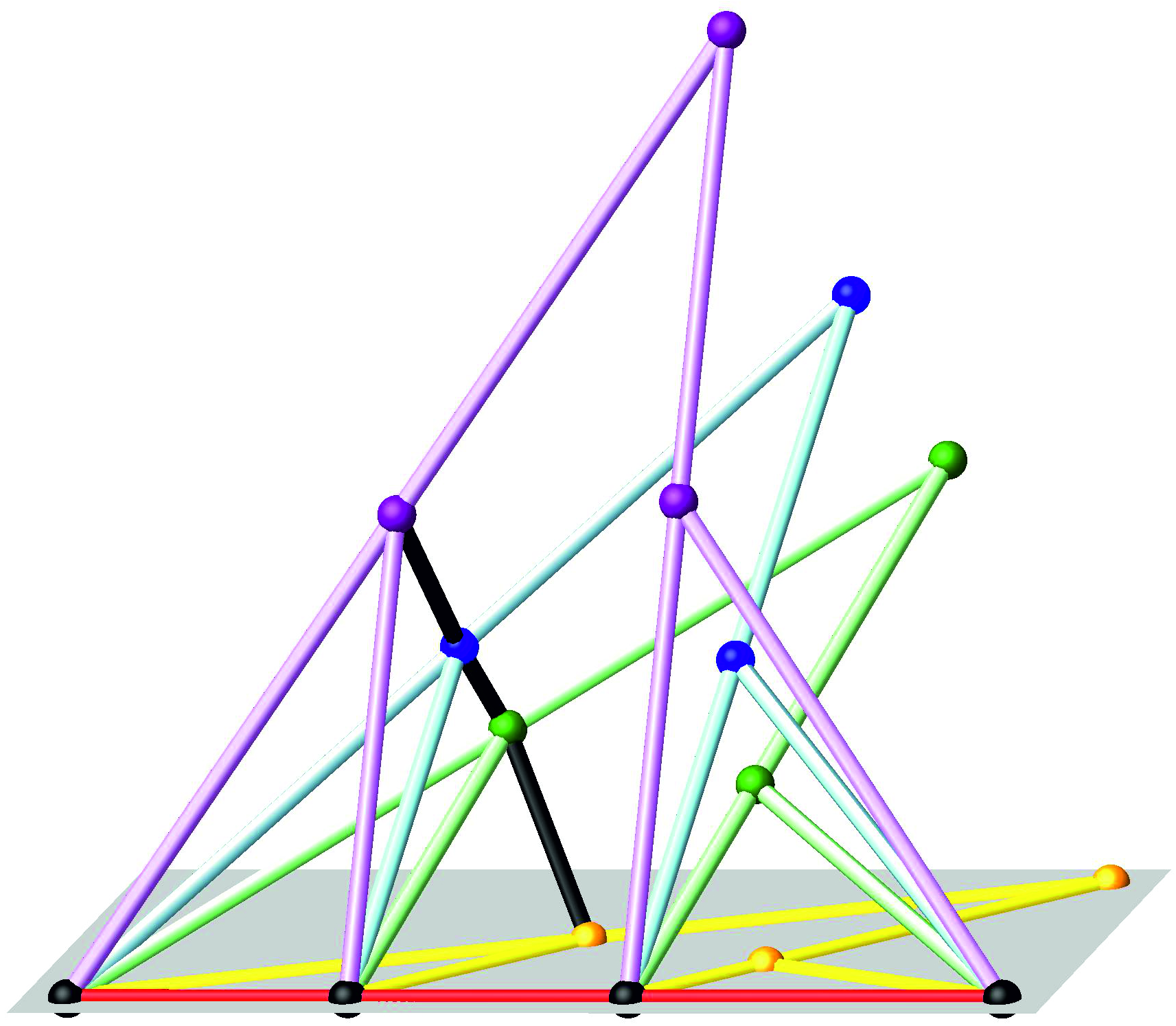}
    \put(38,7){\scriptsize\contour{white} {${V}^{\circ}_{0,0}$}}
    \put(33,22){\scriptsize\contour{white} {${V}^{\circ}_{1,0}$}}
    \put(21.5,41){\scriptsize\contour{white} {${V}^{\circ}_{m,0}$}}
    \put(26,-5.7){\scriptsize\contour{white} {$S_{0}$}}
    \put(2,-5.7){\scriptsize\contour{white} {$S^\pm_{1}$}}
    \put(82,-5.7){\scriptsize\contour{white} {$S_{n}$}}
  \end{overpic} 
\hfill
 \begin{overpic}
    [width=38mm]{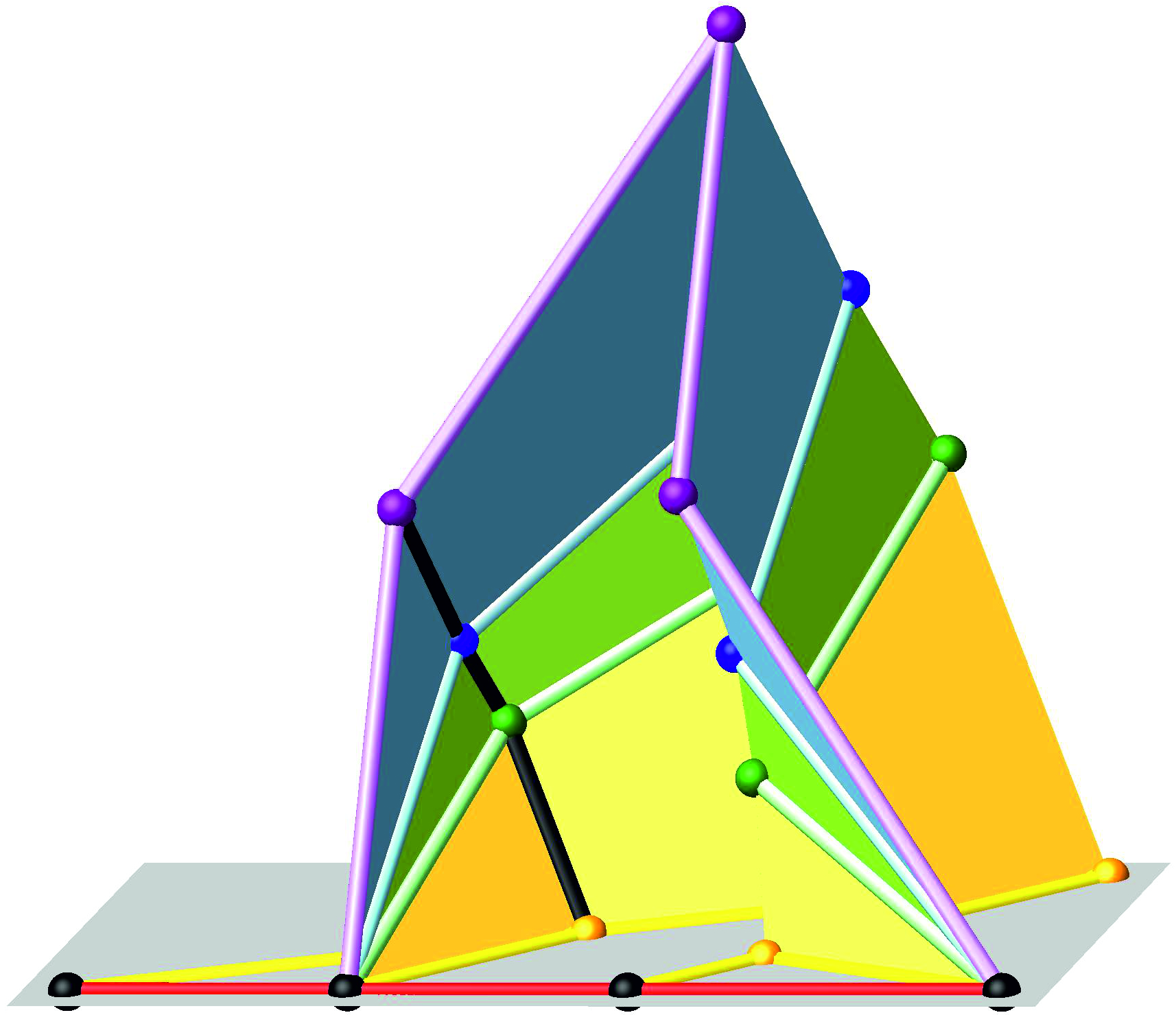}
    \put(38,7){\scriptsize\contour{white} {${V}^{\circ}_{0,0}$}}
    \put(33,21.5){\scriptsize\contour{white} {${V}^{\circ}_{1,0}$}}
    \put(21.5,41){\scriptsize\contour{white} {${V}^{\circ}_{m,0}$}}
    \put(26,-5.7){\scriptsize\contour{white} {$S_{0}$}}
    \put(2,-5.7){\scriptsize\contour{white} {$S^\pm_{1}$}}
    \put(82,-5.7){\scriptsize\contour{white} {$S_{n}$}}
  \end{overpic} 
\hfill	
\begin{overpic}
    [width=38mm]{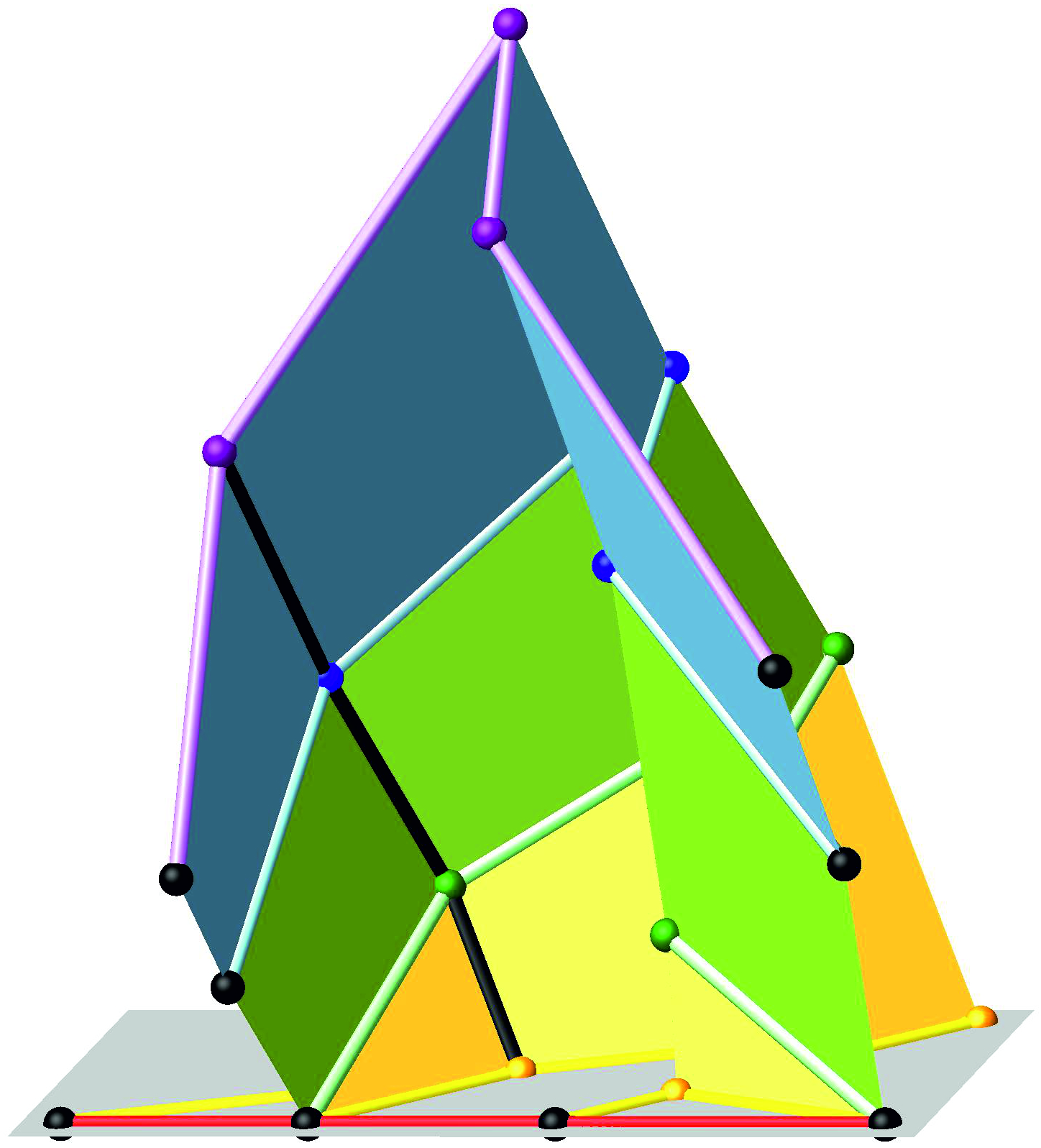}
     \put(34,6.5){\scriptsize\contour{white} {${V}_{0,0}$}}
    \put(28,22.5){\scriptsize\contour{white} {${V}_{1,0}$}}
    \put(8,58){\scriptsize\contour{white} {${V}_{m,0}$}}
    \put(24,-5.0){\scriptsize\contour{white} {$S_{0}$}}
    \put(2,-5.0){\scriptsize\contour{white} {$S^\pm_{1}$}}
    \put(74,-5.0){\scriptsize\contour{white} {$S_{n}$}}
    \put(0,9){\scriptsize\contour{white} {$\tau_2^{-1}(S_{0})$}}
    \put(-10,27){\scriptsize\contour{white} {$\tau_3^{-1}(\tau_2^{-1}(S_{0}))$}}
    \put(72.3,19){\scriptsize\contour{white} {$\tau_2^{-1}(S_{n})$}}
    \put(58,35){\scriptsize\contour{white} {$\tau_3^{-1}(\tau_2^{-1}(S_{n}))$}}
  \end{overpic} 
  \medskip
\hfill
\caption{(left)  Iterative composition of the two possible linear constructions $\sigma^+$ and $\sigma^-$ in $\pi^{\circ}_i$ with $V^{\circ}_{i,0}$ as starting point.  Interpretation of the resulting point set as axial P-hedron (center) and the corresponding general P-hedron (right). 
}
  \label{fig2}
\end{figure}

Using these three input polylines the related general P-hedron can be constructed in the following three steps:
\begin{itemize}
\item [1.] In the first step we compute the input of the axial P-hedron associated with the general one as follows (cf.\ Fig.\ \ref{fig1}-center): 
We apply a translation $\tau_i$ to all points $V_{i,0}, \ldots ,V_{m,0}$ and points $D_i, \ldots , D_m$ in direction 
$\overrightarrow{V_{i-1,0},V_{i,0}}$ in such a way that $\tau_i(\pi_i)$ contains the $z$-axis. By iterating this procedure for $i=2,\ldots, m$ we end up with the polylines $V_{0,0}^{\circ}=V_{0,0}, V_{1,0}^{\circ}=V_{1,0},  V_{2,0}^{\circ},  \ldots ,V_{m,0}^{\circ}$ 
and $D_0^{\circ}=D_0, D_1^{\circ}=D_1,  D_2^{\circ}, \ldots , D_m^{\circ}$ and the pencil of direction planes 
$\pi_0^{\circ}=\pi_0, \pi_1^{\circ}=\pi_1, \pi_2^{\circ}, \ldots, \pi_m^{\circ}$. 
As we only applied translations, the following properties hold true:
\begin{equation}
V_{i,0}^{\circ}V_{i+1,0}^{\circ} \parallel V_{i,0}V_{i+1,0}, \quad
V_{i,0}^{\circ}D_{i}^{\circ} \parallel V_{i,0}D_{i}, \quad
\pi_i^{\circ} \parallel \pi_i. 
\end{equation}
This is the parallelism operation of Sauer and Graf \cite{graf,sauer} mentioned in \cite{nawratil}.
\item [2.] 
In each of the planes $\pi^{\circ}_i$ we proceed with an iterative  composition of two possible linear constructions $\sigma^+$ and $\sigma^-$, respectively, to determine the point sequence $V^{\circ}_{i,0},V^{\circ}_{i,1},\ldots, V^{\circ}_{i,n-1}$ (cf.\ Fig.\ \ref{fig2}-left). The linear mappings $\sigma^\pm$ are defined as follows (cf.\ Fig.\ \ref{fig1}-right) according to \cite{nawratil}:
\begin{itemize}
    \item [$\scriptscriptstyle{(+)}$]
    There exists a central scaling $\sigma^+$ which maps $S_j$ to $S_{j+2}$  with center $S^+_{j+1}$. 
    \item [$\scriptscriptstyle{(-)}$]
    There exists a perspective collineation $\sigma^-$ which maps $S_j$ to $S_{j+2}$ with center $S^-_{j+1}$ and the bisector $q$ of  $S_j$ and $S_{j+2}$ as axis. 
\end{itemize}
Then $V^{\circ}_{i,j+1}$ can be constructed from $V^{\circ}_{i,j}$ as $\sigma^\pm(V^{\circ}_{i,j})$ for $j=0,\ldots, n-2$.
\item [3.] By the end of the second step we already obtain all points of the axial P-hedron  (cf.\ Fig.\ \ref{fig2}-center), where it can also be seen that the $S_i$ are the apexes of cones.

For reconstruction of the general P-hedron we have to apply iteratively the translations $\tau_i^{-1}$ to all points 
$V^{\circ}_{i,0},V^{\circ}_{i,1},\ldots, V^{\circ}_{i,n-1}$, \ldots , 
$V^{\circ}_{m,0},V^{\circ}_{m,1},\ldots, V^{\circ}_{m,n-1}$ for
$i=2,\ldots, m$. 
In order to generate a boundary we also apply this series of transformations to the points $S_0$ and $S_n$, respectively (cf.\ Fig.\ \ref{fig2}-right).
\end{itemize}

\begin{figure}
    \begin{overpic}[width=28.5mm]{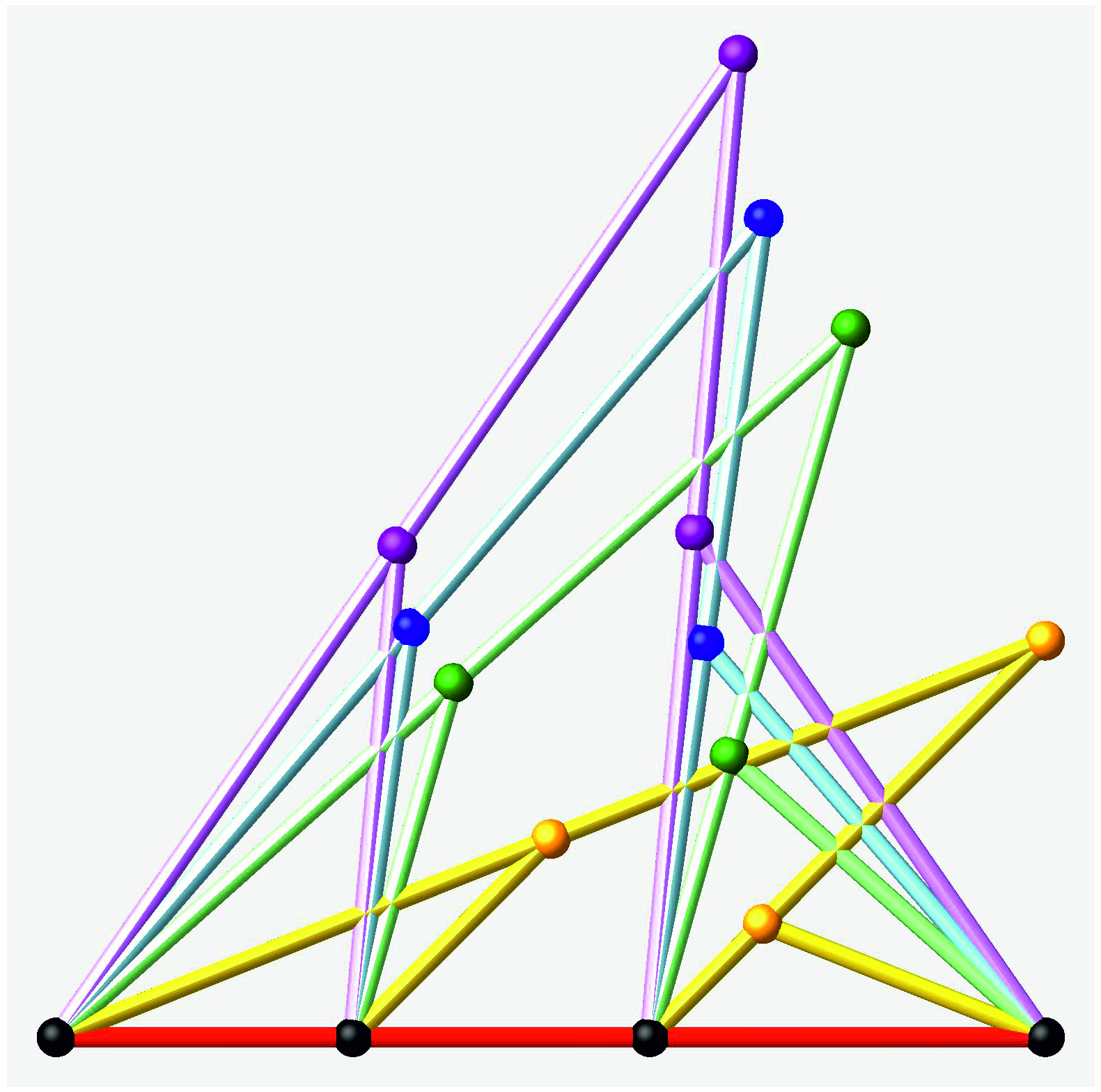}
    \put(24,9){\scriptsize\contour{white} {$S_{0}$}}
    \put(0.5,11){\scriptsize\contour{white} {$S^+_{1}$}}
    \put(51,10){\scriptsize\contour{white} {$S^-_{2}$}}
    \put(92.5,9){\scriptsize\contour{white} {$S_{3}$}}
  \end{overpic}	
  \hfill
   \begin{overpic}[width=28.5mm]{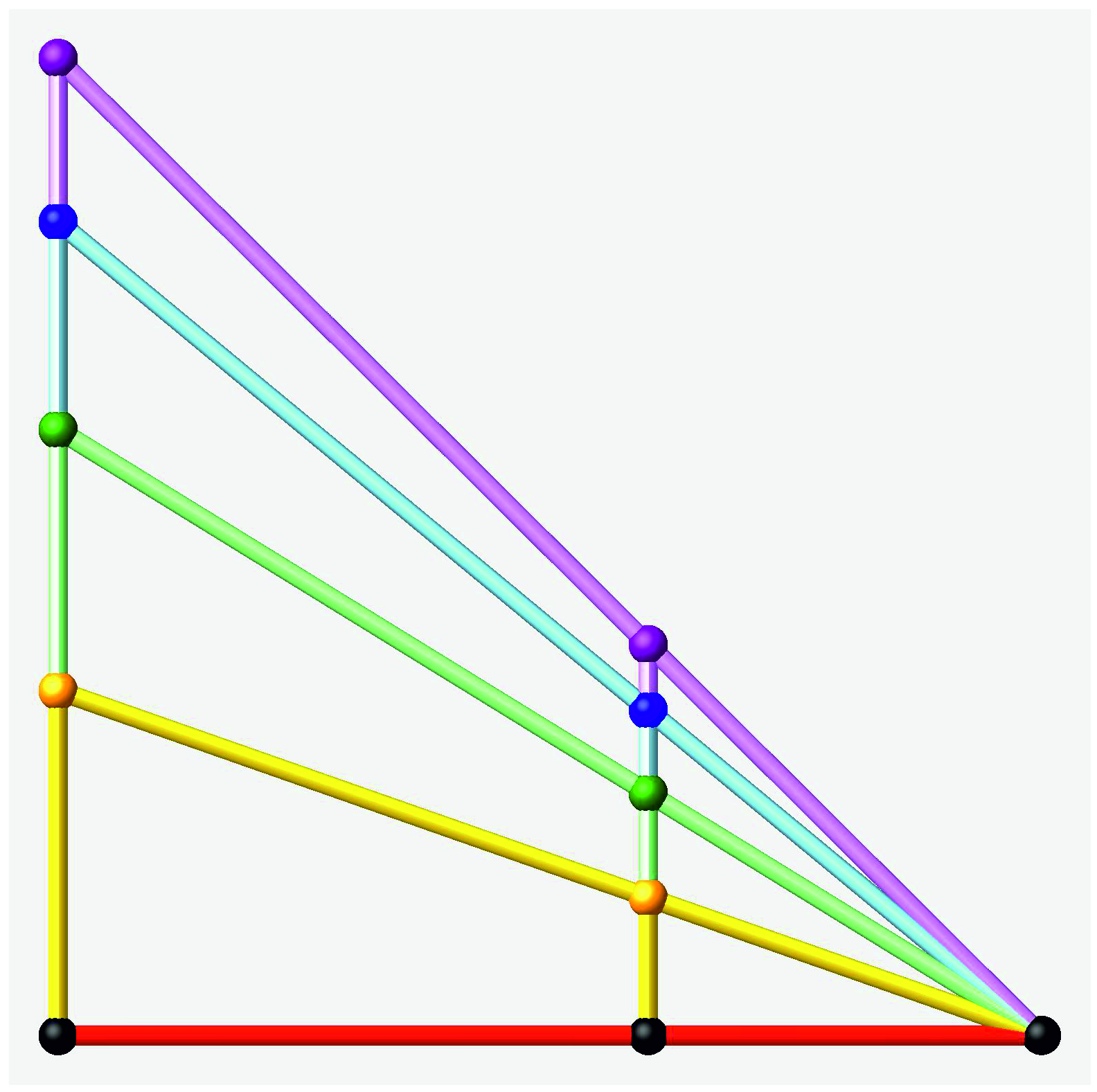}
    \put(7,9){\scriptsize\contour{white} {$S_{0}$}}
    \put(90.5,11){\scriptsize\contour{white} {$S^\pm_{1}$}}
    \put(50,9){\scriptsize\contour{white} {$S_{2}$}}
    \put(9,92){\scriptsize\contour{white} {${V}^{\circ}_{i,0}$}}
    \put(60,43){\scriptsize\contour{white} {$\sigma^{\pm}({V}^{\circ}_{i,0})$}}
  \end{overpic}	
  \hfill
   \begin{overpic}[width=28.5mm]{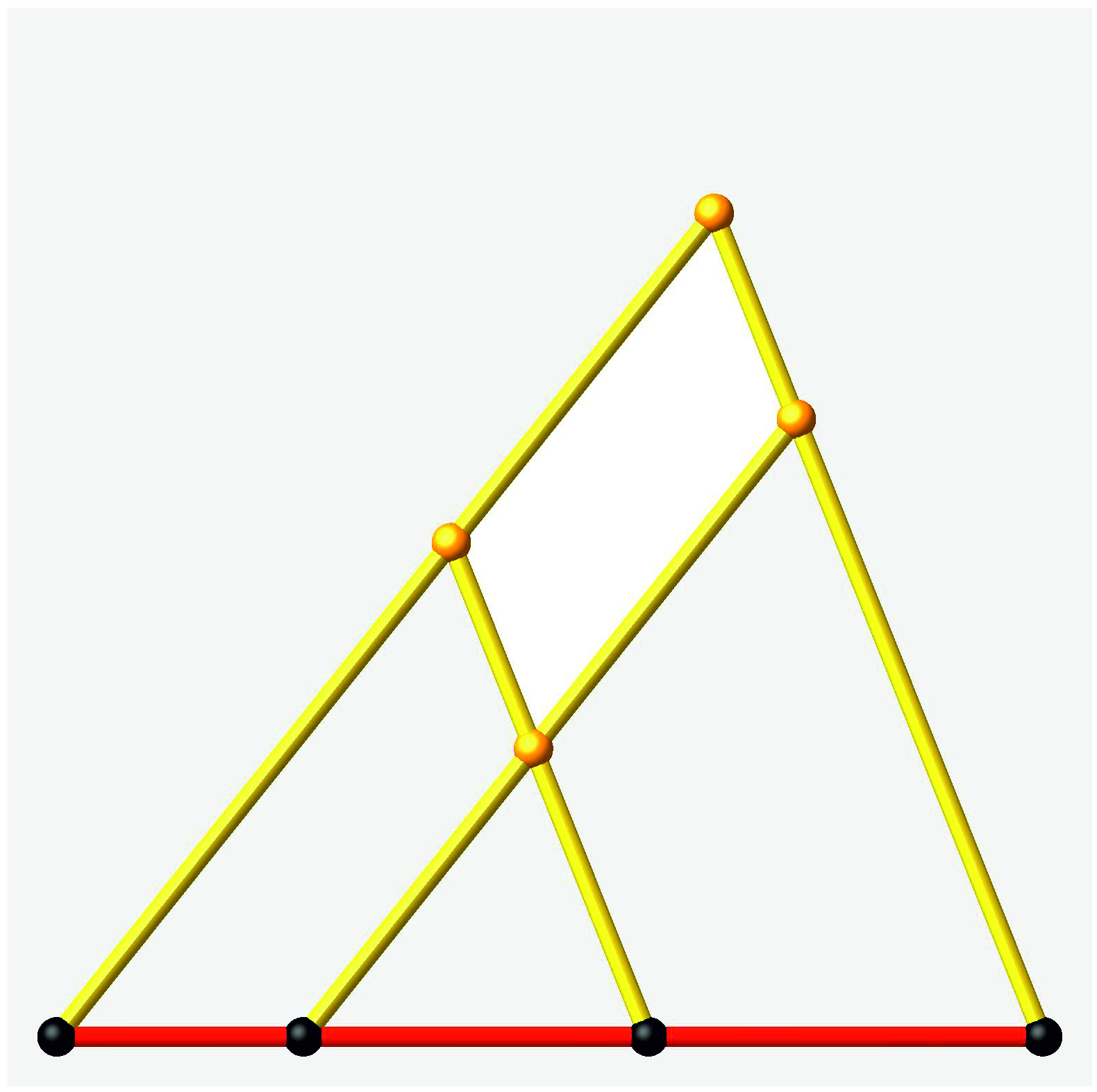}
    \put(28,10){\scriptsize\contour{white} {$S^+_{3}$}}
    \put(0.5,10){\scriptsize\contour{white} {$S_5,S^+_{1}$}}
    \put(51,10){\scriptsize\contour{white} {$S_0,S^+_{4}$}}
    \put(91,11){\scriptsize\contour{white} {$S^+_{2}$}}
    \put(9.5,52){\scriptsize\contour{white} {${V}^{\circ}_{0,4},{V}^{\circ}_{0,0}$}}
    \put(50,82){\scriptsize\contour{white} {${V}^{\circ}_{0,1}$}} 
    \put(74.5,62){\scriptsize\contour{white} {${V}^{\circ}_{0,2}$}} 
    \put(52,26){\scriptsize\contour{white} {${V}^{\circ}_{0,3}$}}
  \end{overpic}	
  \hfill
   \begin{overpic}[width=28.5mm]{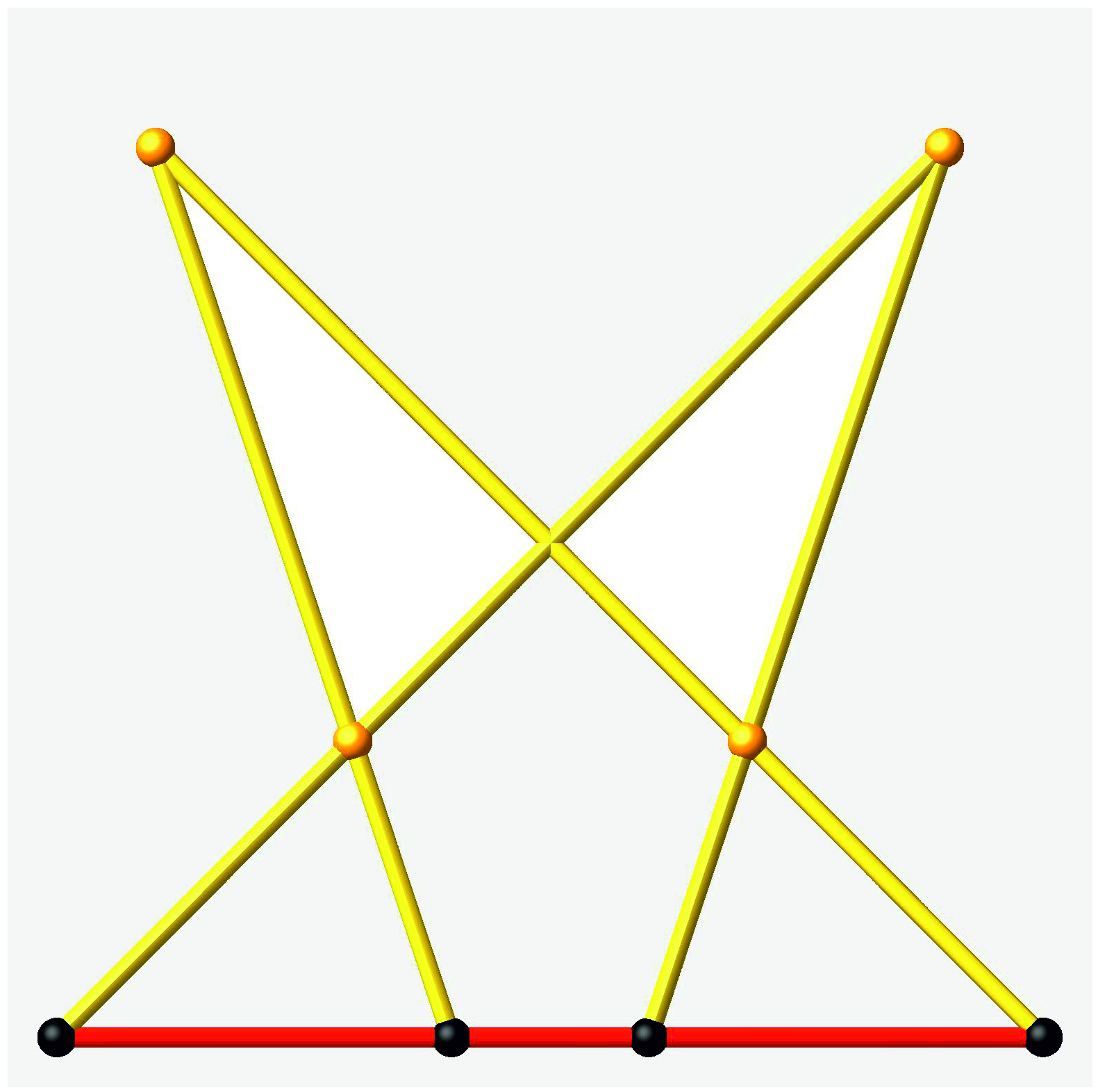}
   \put(58,10){\scriptsize\contour{white} {$S^-_{2}$}}
    \put(0.5,10){\scriptsize\contour{white} {$S_5,S^-_{1}$}}
    \put(30,10){\scriptsize\contour{white} {$S_0,S^-_{4}$}}
    \put(91,11){\scriptsize\contour{white} {$S^-_{3}$}}
    \put(0.5,34){\scriptsize\contour{white} {${V}^{\circ}_{0,4},{V}^{\circ}_{0,0}$}}
    \put(17,87){\scriptsize\contour{white} {${V}^{\circ}_{0,3}$}} 
    \put(71.5,32){\scriptsize\contour{white} {${V}^{\circ}_{0,2}$}} 
    \put(69.5,87){\scriptsize\contour{white} {${V}^{\circ}_{0,1}$}}
  \end{overpic}	
    \caption{(left) The point sets located in the plane $\pi^{\circ}_i$ for $i=0,\ldots m$ can be interpreted as planar linkage $L_i$ with  mobility 1. If we rotate all planes $\pi^{\circ}_i$ for $i=1,\ldots m$ into 
    $\pi^{\circ}_0$, we get the illustrated overconstrained planar linkage $\mathcal{L}$ discussed in \cite{nawratil}. (middle-left) Bifurcation configuration of Remark \ref{rem:no_bifur}. Illustration of the linkage $L_0$ enclosing a parallelogram (middle-right) and an anti-parallelogram (right), respectively.
    } \label{fig3}
\end{figure}
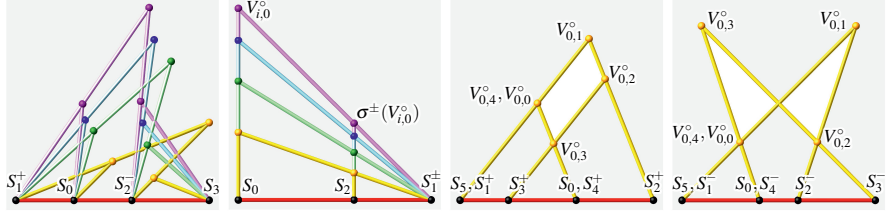

\section{Isometric deformations within the class of P-hedra}\label{sec:defomration}
As a result of the reconstruction done in Section \ref{sec:construction}, we can assume w.l.o.g.\ that the lengths of all edges of the general P-hedron and its associated axial P-hedron are known. As the general P-hedron can by obtained from the axial one by step 3 of Section \ref{sec:construction}, we can restrict ourselves to the parametrization of the isometric deformation of the axial P-hedron, which is explained in the following two subsections.

\subsection{Parametrizing the motion of the linkage $L_0$}
We assumed that $V^{\circ}_{0,0}$ is located on the positive $x$-axis and this property should be kept during the deformation. 
We do not use this $x$-coordinate of $V^{\circ}_{0,0}$ as motion parameter $t$ but the $z$-coordinate of $S_0$ or $S_1$ according to the following criterion: 
\begin{align*}
&\text{Case (a)} &\, &t=z_0 
&\text{for} &\, &\|S_0-V^{\circ}_{0,0}\| <  \|S_1-V^{\circ}_{0,0}\| \\ 
&\text{Case (b)} &\, &t=z_1 
&\text{for} &\,  &\|S_0-V^{\circ}_{0,0}\| >  \|S_1-V^{\circ}_{0,0}\|  
\end{align*}
The reason for this choice is that under a continuous deformation of the configuration  $S_0,V^{\circ}_{0,0},S_1$ only the end point of the shorter bar can switch the sign of its $z$-coordinate. In order to avoid an unnecessary distinction\footnote{For the same reason we do not select the $x$-coordinate of $V^{\circ}_{0,0}$ as motion parameter.} of cases we assume that this $z$-coordinate equals the motion parameter $t$. Note that $t_*$ indicates the time instant of the deformation which corresponds to the initial given configuration.

Until now neither case (a) nor case (b) is covering the possibility $\|S_0-V^{\circ}_{0,0}\| =  \|S_1-V^{\circ}_{0,0}\|$. In this case we consider the smallest $i$ with $\|S_0-V^{\circ}_{i,0}\| \neq  \|S_1-V^{\circ}_{i,0}\|$. For $\|S_0-V^{\circ}_{i,0}\| <  \|S_1-V^{\circ}_{i,0}\|$ we apply case (a); otherwise case (b). The equality $\|S_0-V^{\circ}_{i,0}\| =  \|S_1-V^{\circ}_{i,0}\|$ cannot hold for all $i=0,\ldots ,m$  due to the assumption formulated in the fourth paragraph of Section \ref{sec:construction}.

\begin{remark}\label{rem:sissor}
If $\|S_0-V^{\circ}_{i,0}\| =  \|S_1-V^{\circ}_{i,0}\|$
holds for at least one $i\in\left\{0,\ldots, m\right\}$ then the apexes polyline can be assigned with only pluses\footnote{Note that in this special case $L_i$ is a scissor-like linkage.}; i.e.\ 
$S_0, S_1^+ ,\ldots ,S_{n-1}^+, S_{n}$, because otherwise some of the points $V^{\circ}_{i,1}, \ldots , V^{\circ}_{i,n-1}$ would drop to infinity. This can easily be seen by adopting the mapping $\sigma^-$ illustrated in  Fig.\ \ref{fig1}-right to this special case. \hfill $\diamond$
\end{remark}

After clarifying the choice of the motion parameter $t$ we proceed as follows: 
\begin{enumerate}[{ad} a)]
    \item We start with $S_0(t)$ and then we parametrize $V^{\circ}_{0,0}(t)=(x_{0,0}(t),0,0)^T$ by
        $x_{0,0}(t):=\sqrt{\|V^{\circ}_{0,0}-S_0\|^2-t^2}$. 
    From this we can compute  $S_1(t)=(0,0,z_1(t))^T$ by
        $z_1(t)= \sgn(z_1)\sqrt{\|V^{\circ}_{0,0}-S_1\|^2-x_{0,0}(t)^2}$. 
    \item In this case we start with $S_1(t)$ and parametrize $V^{\circ}_{0,0}(t)=(x_{0,0}(t),0,0)^T$ by
        $x_{0,0}(t):=\sqrt{\|V^{\circ}_{0,0}-S_1\|^2-t^2}$. 
    From this we can compute  $S_0(t)=(0,0,z_0(t))^T$ by
        $z_0(t)= \sgn(z_0)\sqrt{\|V^{\circ}_{0,0}-S_0\|^2-x_{0,0}(t)^2}$.
\end{enumerate}
After this discussion of cases we end up with the three parametrized points $S_0(t)$, $V^{\circ}_{0,0}(t)$, 
$S_1(t)$.  The remainder of this subsection is formulated in a way that it fits for both cases. 

We proceed with the computation of $V^{\circ}_{0,1}(t)=(x_{0,1}(t),0,z_{0,1}(t))^T$ by
\begin{equation}\label{eq:vt}
    V^{\circ}_{0,1}(t)= S_1(t)+\sgn(k)\frac{\|V^{\circ}_{0,1}-S_1\|}{\|V^{\circ}_{0,0}(t)-S_1(t)\|}{\left(V^{\circ}_{0,0}(t)-S_1(t)\right)}
\end{equation}
where ${k:=\langle V^{\circ}_{0,1}-S_1, V^{\circ}_{0,0}-S_1 \rangle}$. 
Then we can compute $S_2(t)$ by
\begin{equation}\label{eq:st}
S_2(t)=V^{\circ}_{0,1}(t)+\sgn(k)\frac
{\|V^{\circ}_{0,1}-S_1\| \|V^{\circ}_{0,0}-S_0\|  }
{\|V^{\circ}_{0,0}-S_1\| \|S_0(t)-V_{0,0}(t)\|}   
M^\pm\left(S_0(t)-V^{\circ}_{0,0}(t)\right)
\end{equation}
where the choice of $M^+=\diag(1,1,1)$ and $M^-=\diag(1,1,-1)$, respectively, depends on the sign $\pm$ associated with $S_1(t)$, which can be assumed w.l.o.g.\ to be the same as assigned to $S_1$ (cf.\ Remark \ref{rem:no_bifur}).  

\begin{remark}\label{rem:no_bifur}
    The sign $\pm$  of $S^\pm_{1}(t)$ can only change under the isometric deformation in a configuration, where $\sigma^+$ and $\sigma^-$ act in the same way on the points 
    $V_{i,0}^{\circ}(t)$ for $i=0,\ldots,m$. Such a bifurcation can only happen in the configuration illustrated in Fig.\ \ref{fig3}-middle-left, but then the P-hedron also belongs to the class of T-hedra (cf.\ fourth paragraph of Section \ref{sec:construction}).  
    Therefore a non-T-hedral P-hedron cannot have a bifurcation configuration implied by a change of the mappings $\sigma^+$ and $\sigma^-$.   \hfill $\diamond$
\end{remark}

An iteration of Eqs.\ (\ref{eq:vt}) and (\ref{eq:st}) by rising the indices implies a parametrization of the complete planar linkage $L_0$; i.e.\ we get $S_0(t),\ldots, S_n(t)$ and 
$V^{\circ}_{0,0}(t), \ldots , V^{\circ}_{0,n-1}(t)$.

\begin{figure}[t]
\begin{overpic}
    [width=120mm]{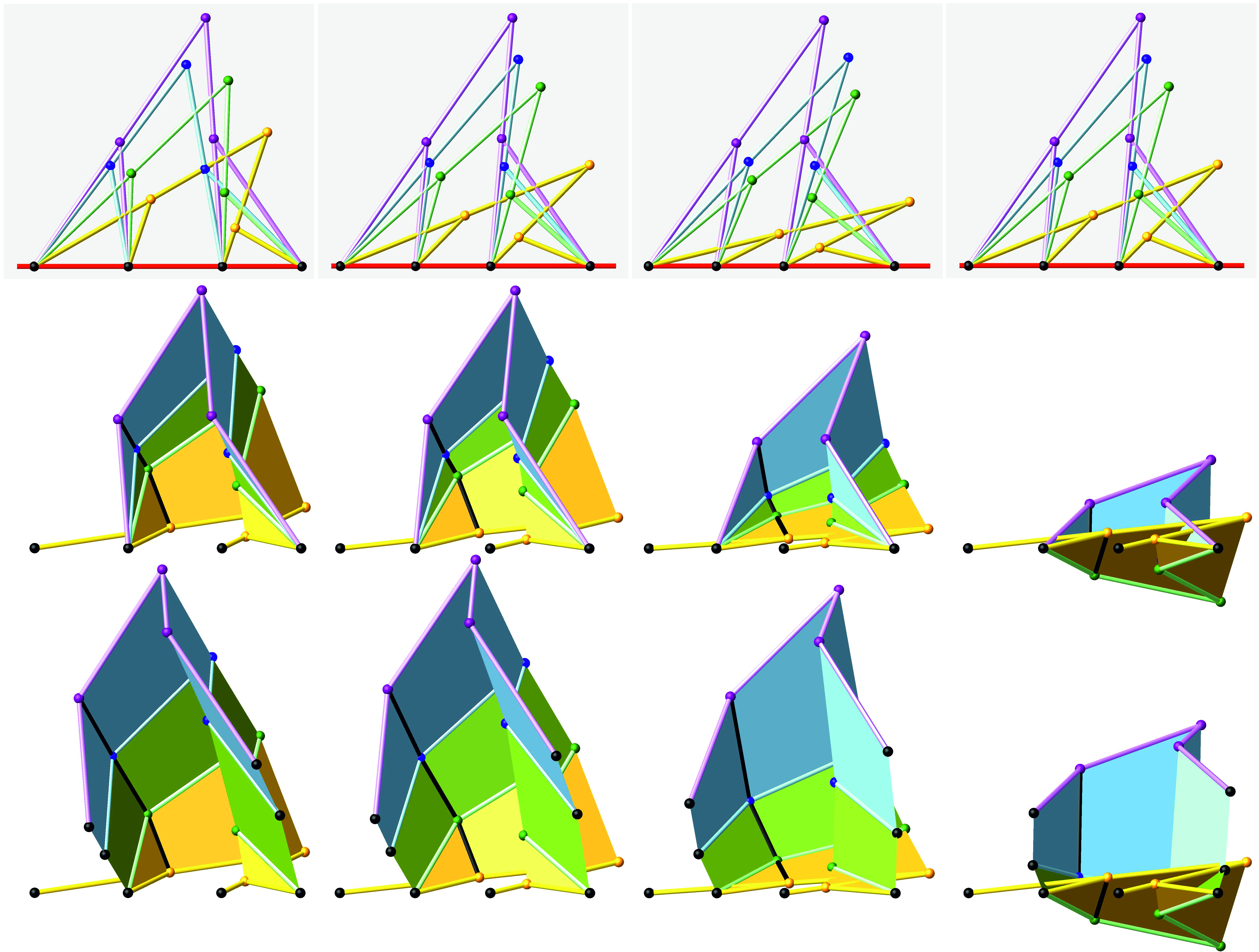}
  \end{overpic}
\caption{Sequence of isometric deformation of the
planar linkage $\mathcal{L}$ (top), 
axial P-hedron (middle) and general P-hedron (bottom). The corresponding animations can be downloaded from \url{https://www.geometrie.tuwien.ac.at/nawratil/publications.html}. The second column corresponds to the given configuration (parameter $t_*$) already illustrated in Figs.\ \ref{fig2} and \ref{fig3}-left. The third column illustrates a flexion limit and the fourth column displays the other branch for the parameter $t_*$. Therefore the planar linkage $\mathcal{L}$ has the same configuration in the second and fourth column. Note that both P-hedra of the fourth column have self-intersections.
}
  \label{fig4}
\end{figure}

\subsection{Parametrizing the motion of the trajectory polyline}

Let us start with the computation of $V^{\circ}_{1,0}(t)=(x_{1,0}(t),y_{1,0}(t),z_{1,0}(t))^T$. The three coordinates can be computed from the following system of equations:
\begin{equation}\label{eq:system}
\begin{split}
\|V^{\circ}_{1,0}(t)-S_0(t)\|^2-  &\|V^{\circ}_{1,0}-S_0\|^2=0,\quad 
\|V^{\circ}_{1,0}(t)-S_1(t)\|^2-  \|V^{\circ}_{1,0}-S_1\|^2=0, \\
&\|V^{\circ}_{1,0}(t)-V_{0,0}(t)\|^2-  \|V^{\circ}_{1,0}-V_{0,0}\|^2=0, 
\end{split}
\end{equation}
having two solutions, which are plane symmetric with respect to $\pi_0^{\circ}$. We denote the branch containing $V^{\circ}_{1,0}$ by  $V^{\circ}_{1,0}(t)$; i.e.\ $V^{\circ}_{1,0}(t_*)=V^{\circ}_{1,0}$; and the other one by $\dzl V^{\circ}_{1,0}(t)$. 

Now we select the branch $V^{\circ}_{1,0}(t)$. Using this parametrization we can compute  $V^{\circ}_{2,0}(t)$ by solving an analogous system to Eqs.\ (\ref{eq:system}). We can iterate this procedure until we get the  complete parametrized trajectory polyline $V^{\circ}_{0,0}(t),V^{\circ}_{1,0}(t), \ldots,  V^{\circ}_{m,0}(t)$. 

\begin{remark}\label{rem:bifur}
Common configurations of the two branches $V^{\circ}_{i,0}(t)$ and  $\dzl V^{\circ}_{i,0}(t)$ are characterized by the coplanarity of the involved points (i.e.\ the profile planes $\pi^{\circ}_{i-1}$ and $\pi^{\circ}_{i}$ coincide\footnote{This characterizes also the flexion limits/bifurcation configurations of T-hedra (cf.\ \cite{graf,SNRT2021,sauer}).}); which correspond to zeros of 
$\det(V^{\circ}_{i,0}(t)-S_0(t),V^{\circ}_{i,0}(t)-S_1(t),V^{\circ}_{i,0}(t)-V_{i-1,0}(t))=0$. If one computes all real zeros for $i=1,\ldots,m$ and sort them together with $t_*$ we obtain the sequence:
$t_\alpha\leq t_\beta\leq\ldots\leq t_\lambda < t_* < t_\mu\leq \ldots\leq t_\omega$. This also implies the flexion interval $t\in [t_\lambda,t_\mu]$ of the P-hedron, because by overshooting the values $t_\lambda$ and $t_\mu$ the solutions of the corresponding system (\ref{eq:system}) turn from real to complex. Note that due to the assumption related to Footnote \ref{fn:bifur} the length of the flexion interval cannot be zero; i.e.\ there always exists a real flex out of the given configuration of the P-hedron. We can follow this isometric deformation until the  flexion limits $t_\lambda$ and $t_\mu$ are reached. As they are also bifurcation configurations we can switch over to the other branch and flex back (cf.\ Fig.\ \ref{fig4}). \hfill $\diamond$
\end{remark}

The parametrization of the remaining vertices of the axial P-hedron can be completed by using $V^{\circ}_{i,0}(t)$ as starting point for the iteration of the analogous equations to (\ref{eq:vt}). In this way we get the points
$V^{\circ}_{i,1}(t), \ldots , V^{\circ}_{i,n-1}(t)$ for $i=1,\ldots ,m$.

A test implementation of the given algorithm was done for verification in Maple, which was also used to produce the example illustrated in Fig.\ \ref{fig4}. For real-time interactive handling it is planed to implement this algorithm\footnote{The special case mentioned in Footnote \ref{fn:special} has to be coded separately.} in a Rhino/Grasshopper plugin.  
Such a plugin can also be used to approximate the isometric deformation of semi-discrete P-hedra, by discretizing a given smooth input trajectory into a polyline with sufficiently small line-segments.

\section{Final remarks}\label{sec:final}

{\bf $\bullet$ Developable pattern} play an important role in origami and fabrication. 
Developable (but also flat-foldable)  pattern can be considered as special P-hedra where all bifurcation possibilities arise at the same time; i.e.\ all $\pi_i$ collapse into one plane in a configuration. Clearly one can construct the P-hedron in this developed (flat-folded) configuration for which $V_{i,j}=V^{\circ}_{i,j}$ and $\pi_i=\pi^{\circ}_i$ hold true. But now one cannot be sure that a real flex out of the constructed configuration exists, as the relation $t_\lambda = t_* = t_\mu$ (in terms of Remark \ref{rem:bifur}) could hold true. 

For the existence of a real flex we only have to show that the planar linkage $\mathcal{L}$ (cf.\ Fig.\ \ref{fig3}) can be associated with an infinitesimal motion ($\neq$ instantaneous standstill) in a way that the distances between 
$V^{\circ}_{i,j}$ and $V^{\circ}_{i+1,j}$ for all $i=0,\ldots, m-1$ and $j=0,\ldots, n-1$ do not expand instantaneously, as otherwise the  PQ-mesh would tear apart. This non-expansion can easily be checked by the following criterion:
\begin{equation}
    \langle v(V^{\circ}_{i,j}),V^{\circ}_{i,j}-V^{\circ}_{i+1,j}\rangle
    +  \langle v(V^{\circ}_{i+1,j}),V^{\circ}_{i+1,j}-V^{\circ}_{i,j}\rangle \leq 0
\end{equation}
where $v(.)$ denotes the velocity of the point associated with the planar linkage $\mathcal{L}$. The determination of these velocities is a standard procedure in the kinematics of planar mechanisms, which can even be done in a pure graphical way (e.g.\ \cite{wunderlich}).  

\begin{remark}
    Note that developable/flat-foldable P-hedra belong to the class of
    conic equimodular PQ-surfaces in the classification of Izmestiev \cite{Izm17}. Their semi-discrete analogues belong to the nets discussed in \cite{klara}. \hfill $\diamond$
\end{remark}

\noindent
{\bf $\bullet$ P-hedral tubes} are further interesting objects, as they can be regarded as building blocks of rigid-foldable meta-materials/surfaces (cf.\ \cite{KMN1}). They are obtained by constructing a linkage $L_0$ in a way that $V^{\circ}_{0,0}(t)= V^{\circ}_{0,n-1}(t)$
holds true for all $t\in [t_\lambda,t_\mu]$. For $n=5$ this can only be the case\footnote{Both cases also appear for T-hedra, where in  addition the deltoidal case exists (cf.\ \cite{KMN1}).} for a parallelogram or an anti-parallelogram, where the symmetry line of latter one has to be orthogonal to the axis of the axial P-hedron (see Fig.\ \ref{fig3}). The parallelogram results in a rigid-foldable prismatic tube and the anti-parallelogram in a composition of plane-symmetric Bricard octahedra. 

In analogy to \cite{KMN1} we can also combine these flexible tubes by edge-sharing and/or the aligned-coupling of faces. By deleting the common faces of the latter coupling one can produce more complicated P-hedral tubes than the two mentioned above. \medskip

\noindent
{\bf $\bullet$ Future research} is dedicated to the study of (i) zipper couplings of P-hedral tubes (cf.\ \cite{KMN1}) and (ii) isometric deformations of translational surfaces contained in the class of general P-hedra, which were mentioned in Footnote \ref{fn:trans}.

\begin{acknowledgement}
This research was funded in whole or in part by the Austrian Science Fund (FWF) [grant F77: SFB ``Advanced Computational Design'', subproject 7]. For open access purposes, the author has applied a CC BY public copyright license to any author accepted manuscript version arising from this submission.
\end{acknowledgement}


\begin{thebibliography}{99.}

\bibitem{Izm17}	
Izmestiev, I.: Classification of flexible Kokotsakis polyhedra with quadrangular base. 
International Mathematics Research Notices \textbf{2017}(3):715--808 (2017) 

\bibitem{voss}
Kilian, M., Nawratil, G., Raffaelli, M., Rasoulzadeha, A., Sharifmoghaddama, K.: 
Interactive design of discrete Voss nets and simulation of their rigid foldings. Computer Aided Geometric Design \textbf{111}:102346 (2024)

\bibitem{graf}
Sauer, R., Graf, H.: \"Uber Fl\"achenverbiegung in Analogie zur Verknickung offener Facettenflache. 
Mathematische Annalen \textbf{105}:499--535 (1931)

\bibitem{SNRT2021}
Sharifmoghaddam, K., Nawratil, G., Rasoulzadeh, A., Tervooren, J.: Using Flexible Trapezoidal Quad-Surfaces for Transformable Design. 
Proc. of IASS Annual Symposia, Surrey Symposium: Transformable structures (IASS WG 15), pp.\ 1--13, IASS (2021)

\bibitem{KMN1}
Sharifmoghaddam, K., Maleczek, R., Nawratil, G.:
Generalizing rigid-foldable tubular structures of T-hedral type. Mechanics Research Communications \textbf{132}:104151 (2023)

\bibitem{sauer}
Sauer, R.: Differenzengeometrie. Springer (1970)

\bibitem{arvin}
Izmestiev, I., Rasoulzadeh, A., Tervooren, J.:
Isometric Deformations of Discrete and Smooth {T}-surfaces. Computational Geometry \textbf{122}:102104 (2024)

\bibitem{nawratil}
Nawratil, G.: From axial C-hedra to general P-nets.  Advances in Robot Kinematics (J. Lenarcic, M. Husty eds.), pages 340--347, Springer, 2024

\bibitem{wunderlich}
Wunderlich, W.: Ebene Kinematik. 
Bibliographisches Institut AG, Mannheim (1970)

\bibitem{klara}
Mundilova, K.: On the rigid-ruling folding of curved creases: Conjugate-net preserving isometric deformations of semi-discrete globally developable conjugate-nets. preprint (2025)

\end{thebibliography}
\end{document}